\documentclass[sigconf]{acmart}

\AtBeginDocument{%
  \providecommand\BibTeX{{%
    \normalfont B\kern-0.5em{\scshape i\kern-0.25em b}\kern-0.8em\TeX}}}

\copyrightyear{2021}
\acmYear{2021}
\setcopyright{acmlicensed}\acmConference[SIGIR '21]{Proceedings of the 44th International ACM SIGIR Conference on Research and Development in Information Retrieval}{July 11--15, 2021}{Virtual Event, Canada}
\acmBooktitle{Proceedings of the 44th International ACM SIGIR Conference on Research and Development in Information Retrieval (SIGIR '21), July 11--15, 2021, Virtual Event, Canada}
\acmPrice{15.00}
\acmDOI{10.1145/3404835.3462828}
\acmISBN{978-1-4503-8037-9/21/07}

\usepackage{booktabs}
\usepackage{threeparttable}
\usepackage{float}
\usepackage{multirow}
\usepackage{textcomp}
\usepackage{caption}
\usepackage{booktabs}
\usepackage{bm}
\usepackage{makecell}
\usepackage{url}
\usepackage[labelformat=simple]{subcaption}
\usepackage{CJKutf8}

\newcommand{\bE}{\mathbf{E}}

\newcommand{\bR}{\mathbf{R}}

\newcommand{\bc}{\mathbf{c}}
\newcommand{\bd}{\mathbf{d}}
\newcommand{\be}{\mathbf{e}}

\newcommand{\bh}{\mathbf{h}}

\newcommand{\bk}{\mathbf{k}}

\newcommand{\bp}{\mathbf{p}}

\newcommand{\br}{\mathbf{r}}

\newcommand{\bv}{\mathbf{v}}

\newcommand{\bx}{\mathbf{x}}
\newcommand{\by}{\mathbf{y}}

\newcommand{\ie}{\textit{i.e.}}
\newcommand{\eg}{\textit{e.g.}}

\begin{document}
\fancyhead{}

\title{One Chatbot Per Person: Creating Personalized Chatbots based on Implicit User Profiles 
}

\author{Zhengyi Ma$^{2,3}$, Zhicheng Dou$^{1,3}$, Yutao Zhu$^{5}$, Hanxun Zhong$^{2,3}$, and Ji-Rong Wen$^{3,4}$}

\affiliation{$^1$ Gaoling School of Artificial Intelligence, Renmin University of China
}

\affiliation{$^2$ School of Information, Renmin University of China}

\affiliation{$^3$ Beijing Key Laboratory of Big Data Management and Analysis Methods }

\affiliation{$^4$ Key Laboratory of Data Engineering and Knowledge Engineering, MOE}

\affiliation{$^5$ Université de Montréal, Montréal, Québec, Canada}

\email{{zymaa,dou,hanxun_zhong}@ruc.edu.cn, yutao.zhu@umontreal.ca, jirong.wen@gmail.com}

\renewcommand{\shortauthors}{Ma and Dou, et al.}
\renewcommand{\authors}{Zhengyi Ma, Zhicheng Dou, Yutao Zhu, Hanxun Zhong, and Ji-Rong Wen}

\begin{abstract}
Personalized chatbots focus on endowing chatbots with a consistent personality to behave like real users, give more informative responses, and further act as personal assistants. Existing personalized approaches tried to incorporate several text descriptions as explicit user profiles. However, the acquisition of such explicit profiles is expensive and time-consuming, thus being impractical for large-scale real-world applications. Moreover, the restricted predefined profile neglects the language behavior of a real user and cannot be automatically updated together with the change of user interests. In this paper, we propose to learn implicit user profiles automatically from large-scale user dialogue history for building personalized chatbots. Specifically, leveraging the benefits of Transformer on language understanding, we train a personalized language model to construct a general user profile from the user's historical responses. To highlight the relevant historical responses to the input post, we further establish a key-value memory network of historical post-response pairs, and build a dynamic post-aware user profile. The dynamic profile mainly describes what and how the user has responded to similar posts in history. To explicitly utilize users' frequently used words, we design a personalized decoder to fuse two decoding strategies, including generating a word from the generic vocabulary and copying one word from the user's personalized vocabulary. Experiments on two real-world datasets show the significant improvement of our model compared with existing methods. Our code is available at \url{https://github.com/zhengyima/DHAP}

\end{abstract}
\maketitle
%

\section{Introduction}  \label{section:Introduction}

\begin{figure}
\centering
\includegraphics[width=\linewidth]{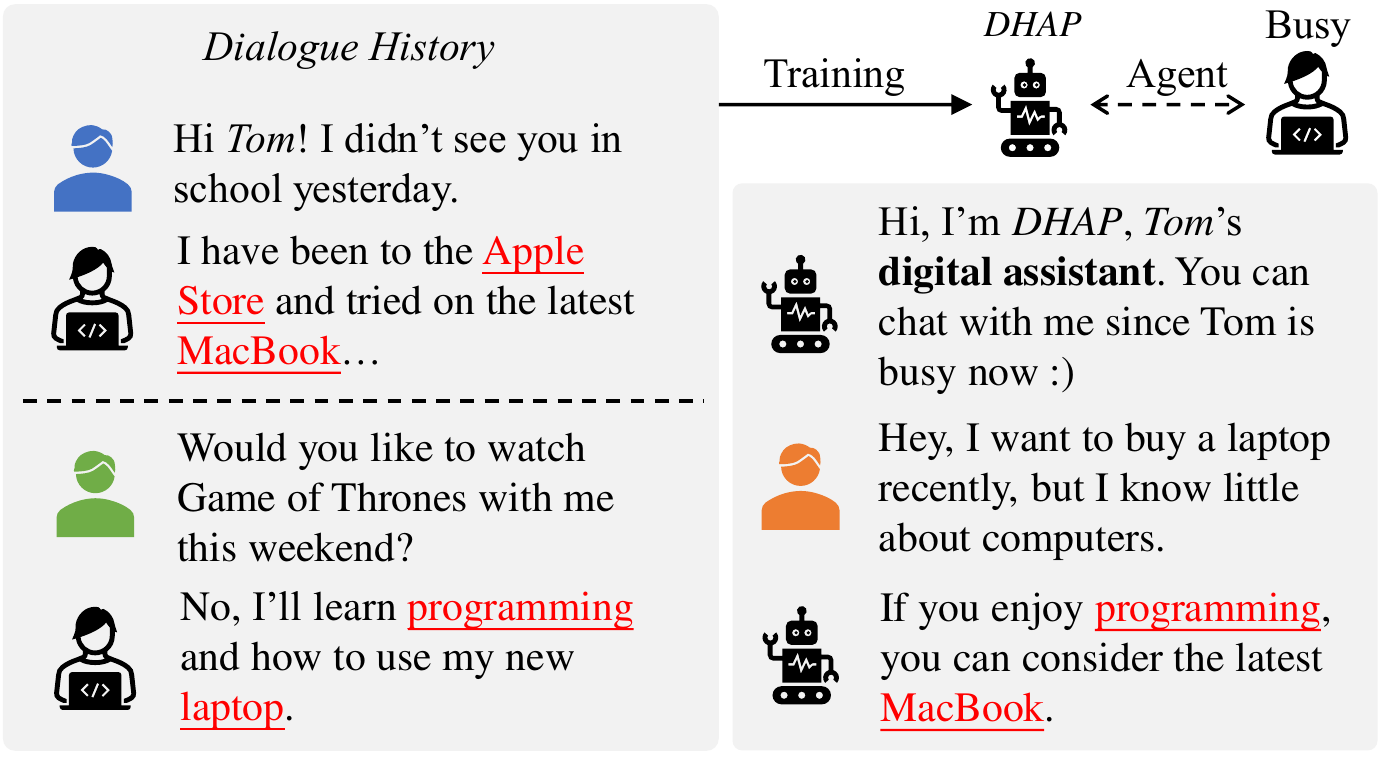}
\caption{An example of the personalized chatbot serving as an intelligent agent for user Tom when Tom is busy. }
\label{fig:intro_example}
\end{figure}

Faced with extensive information available on the Internet, it is very appealing to have an intelligent assistant that can provide the most relevant information~\cite{DBLP:conf/cikm/MaDBW20,DBLP:conf/sigir/ZhouDW20,DBLP:conf/wsdm/ZhouDW20}, collaborate with us on day-to-day problems~\cite{DBLP:conf/aaai/LuoHZN019,DBLP:journals/corr/abs-1812-08989}, or even act as our agent for some specific tasks~\cite{DBLP:journals/ker/WooldridgeJ95}. 
Towards this ultimate goal, in the dialogue system area, building digital agents has attracted more and more attention and had some preliminary applications in our daily life~\cite{DBLP:conf/aaai/LuoHZN019,DBLP:conf/acl/QiuLWGCZCHC17,DBLP:journals/corr/abs-1812-08989}.
In the future, some chit-chat between humans will inevitably be completed by digital agents. In this paper, we carry out a preliminary study toward this goal and focus on the problem of developing personalized chatbots. 
A personalized chatbot aims at leveraging personalized information (\eg, a predefined persona~\cite{DBLP:conf/acl/KielaWZDUS18,DBLP:conf/acl/LiuCCLCZZ20}) to provide personalized responses when communicating with others. 
Such personalized information can help chatbots generate more consistent and informative replies.
More importantly, if the personalized information can be well captured, the chatbots can perform similar behaviors like real users (\eg, serving as an agent of the user and give similar responses to others when the user is busy, as shown in Figure~\ref{fig:intro_example}), thus having the potential to be an intelligent agent with that specific personality~\cite{DBLP:journals/ker/WooldridgeJ95}.

Many models have been proposed for improving the chatbot's capability to generate personalized responses. 
Early studies tried to integrate the user ID embeddings to a sequence-to-sequence (Seq2Seq) model for identifying the user and generating user-related responses~\cite{DBLP:conf/emnlp/ChanLYCHZY19,DBLP:conf/acl/LiGBSGD16,DBLP:conf/emnlp/BakO19}.
Recently, some studies proposed to assign predefined personas to chatbots so as to generate more personalized responses~\cite{DBLP:conf/acl/KielaWZDUS18,DBLP:conf/ijcai/QianHZXZ18,DBLP:conf/aaai/SongZH020}. 
They assumed that personality can be described in several sentences or attributes and help chatbots generate more persona-related responses.
Different from existing studies, in this work, we propose letting the chatbot \textbf{learn the implicit user profile automatically from the user dialogue history} and \textbf{generate personalized responses based on the learned user profile}. 
In this way, the chatbot can be personalized by user's (\eg, user A) historical data, behave like this user, act as this user's agent, and chat with any other users (\eg, user B, C, etc.).

Our idea is motivated by: (1) Contrary to explicit persona descriptions, user dialogue history is easier to be gathered on the user's client devices. 
It is evident that obtaining explicit descriptions for massive users is impractical in real applications~\cite{DBLP:conf/emnlp/ChanLYCHZY19}:
First, users may be lazy to set their profile before using the chatbot~\cite{DBLP:journals/misq/GerlachK91}. 
Second, manually collecting user profiles is costly and time-consuming.
Furthermore, even if the user profile is collected, it cannot be updated with the change of user interests, thus may be ineffective over time. 
Finally, a fixed set of properties is not suitable for describing all users. 
(2) The user dialogue history contains massive personalized information, which is suitable for learning user profiles automatically. 
As shown in Figure~\ref{fig:intro_example}, the dialogue history of a user includes their historical responses, and the corresponding posts
issued by other users. 
Intuitively, users' historical responses can often reflect their language style, background knowledge, frequently used words, and even their interests. 
For example, there may be many electronic device names that appeared in the historical responses of an electronic hobbyist (\eg, \textit{MacBook} in Figure~\ref{fig:intro_example}). 
Besides, the interaction content and style between a specific user and others can be captured from the historical post-response pairs. 
When faced with a new input post, the chatbot can look for the historical data, check how the user has responded to a similar post before, and apply similar interactions to generate a suitable response. In addition, user profiles learned from historical data can be gradually updated with more data being collected. 
In summary, the user dialogue history is easy to obtain and appropriate for building user profile.

To achieve our idea, we propose a model \textbf{DHAP} for personalized chatbots, which focuses on learning implicit user profile from user \textbf{D}ialogue \textbf{H}istory \textbf{A}utomatically and generating \textbf{P}ersonalized responses. In our model, a general user profile representation is \textbf{firstly} constructed from the user's historical responses to capture the general information, including user interest, background knowledge, and speaking style. This is implemented by a personalized language model based on Transformer. \textbf{Then}, we design a personalized post encoder to construct the personalized post representation. The general user profile is utilized in the post encoder to better capture the semantic information of the input post. \textbf{Next}, we build a key-value memory network to store the user's historical post-response pairs. Based on this history memory, the dynamic post-aware user profile is built by highlighting the historical responses relevant to the current input post. \textbf{Finally}, we design a personalized decoder to fuse the learned user profile into the response generation process. The personalized decoder can switch between generating a word from a generic vocabulary and copying a word from the user's personalized vocabulary. Experimental results on two large-scale datasets show that our proposed DHAP significantly outperforms existing response generation models in various evaluation metrics.

Our contributions are three-fold: (1) We learn the implicit user profile from user's dialogue history automatically for generating personalized responses. By this means, our method can be applied without additional annotations on user profiles, and create personalized chatbots as user's digital agents.
(2) We build two kinds of user profiles from the dialogue history, including the general user profile reflecting the user's general information and the dynamic post-aware user profile to apply similar interactions in the historical data for the current input.
(3) We design a personalized decoder to coordinate two personalized decoding strategies, including generating a word from the generic vocabulary and copying a word from the personalized vocabulary to leverage the user's word preference.


\begin{figure*}
\centering
    \includegraphics[width=0.8\textwidth]{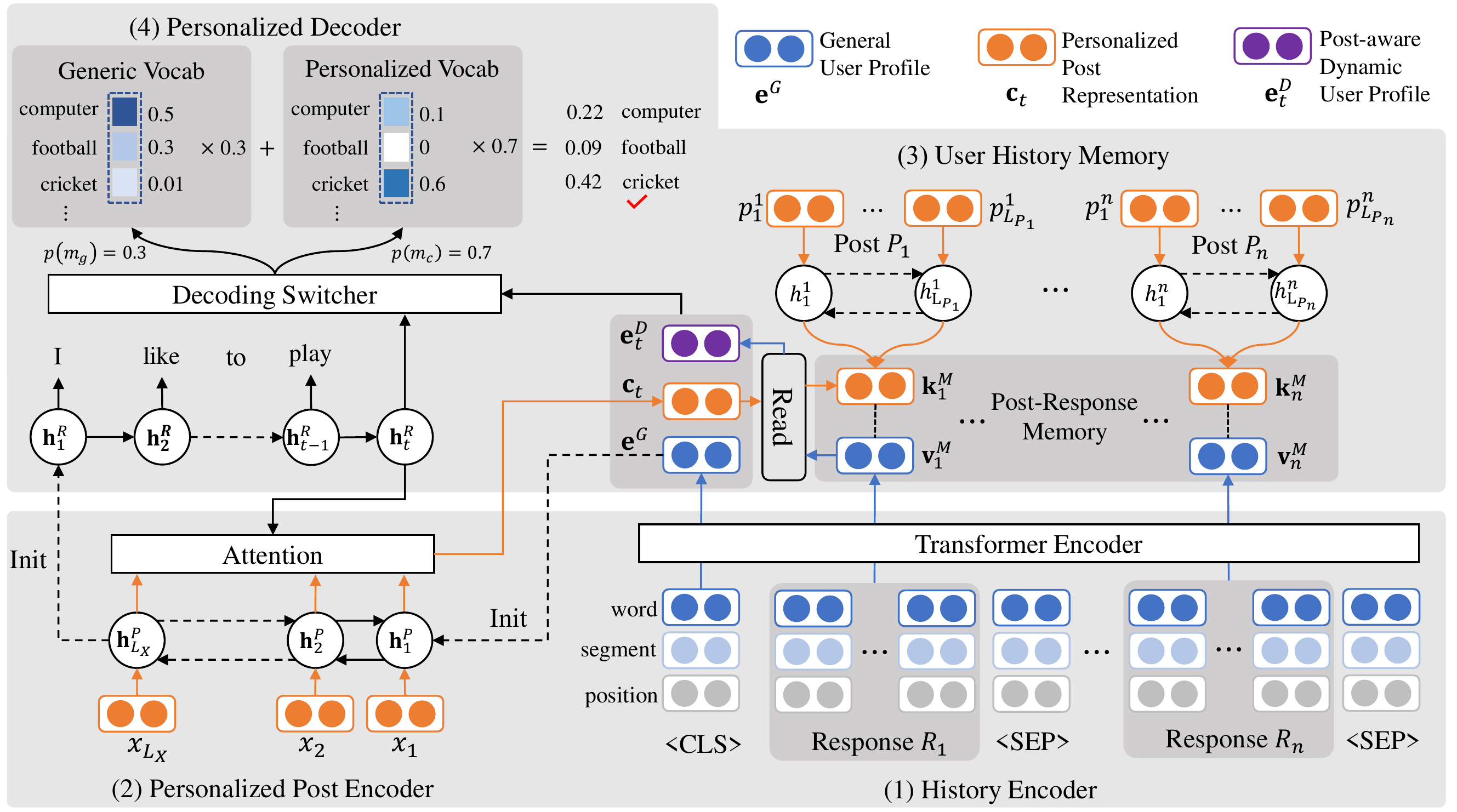}
    \caption{The overall structure of the proposed model DHAP, which consists of (1) a history encoder, (2) a personalized post encoder, (3) a user history memory, and (4) a personalized decoder.}
\label{model_figure}
\end{figure*}

\section{Related Work}\label{sec:related_work}
\paragraph{Open-domain Chatbots and Response Generation}
Open-domain chatbots have attracted more and more attention, due to their broad application in real applications, such as Microsoft XiaoIce~\cite{DBLP:journals/corr/abs-1812-08989}. 
Typical methods can be categorized into two groups: retrieval-based and generation-based. Retrieval-based methods aim to select a suitable response from a large repository~\cite{DBLP:conf/acl/WuWXZL17,DBLP:conf/acl/ZhuSDNZ20,DBLP:conf/ecir/ZhuNZDD21}, while generation-based methods aim at generating a response from scratch~\cite{DBLP:conf/acl/ShangLL15,DBLP:conf/naacl/SordoniGABJMNGD15,DBLP:conf/aaai/SerbanSBCP16,DBLP:journals/ir/ZhuDNW20}. In this study, we focus on the response generation problem. 

Some early studies treat the response generation task as a statistical machine translation problem because of its end-to-end and data-driven features~\cite{DBLP:conf/emnlp/RitterCD11, DBLP:journals/corr/VinyalsL15}. More recently, with the progress of deep learning, Seq2Seq methods have been applied to response generation and achieve great performance~\cite{DBLP:conf/aaai/SerbanSBCP16,DBLP:journals/corr/VinyalsL15,DBLP:conf/aaai/SerbanSLCPCB17, DBLP:conf/naacl/LiGBGD16}. 
Many Seq2Seq-based extensions have been applied to tackle the ``safe response'' problem~\cite{DBLP:conf/naacl/LiGBGD16}; to incorporate external knowledge~\cite{DBLP:conf/ijcai/ZhouYHZXZ18}; to generate responses with emotions or personas~\cite{DBLP:conf/aaai/ZhouHZZL18,DBLP:conf/acl/LiGBSGD16,DBLP:conf/ijcai/QianHZXZ18}; and to model the hierarchical structure of the dialogue context~\cite{DBLP:conf/aaai/SerbanSBCP16,DBLP:conf/aaai/SerbanSLCPCB17}.

\paragraph{Personalized Chatbots}
Endowing chatbots with a coherent personality is a challenging but necessary step to achieve the ultimate goal of building an intelligent assistant. With personality, a chatbot can generate more informative and user-specific responses~\cite{DBLP:conf/acl/LiGBSGD16,DBLP:conf/acl/KielaWZDUS18}, and has the potential to perform similar behaviors as real humans.

Traditional personalized chatbots focus on modeling the user's psychological behavior such as the ``Big Five'' of speakers~\cite{DBLP:conf/acl/MairesseW07}. 
In recent years, deep-learning based methods were proposed to learn the persona information directly from large-scale dialogue datasets via end-to-end neural networks~\cite{DBLP:conf/acl/LiGBSGD16,DBLP:conf/emnlp/ChanLYCHZY19,DBLP:conf/acl/KielaWZDUS18,DBLP:conf/ijcai/QianHZXZ18}. Some researchers first tried to input the user ID embeddings into the decoder of a Seq2Seq model to generate more personalized responses~\cite{DBLP:conf/acl/LiGBSGD16,DBLP:journals/corr/Al-RfouPSSSK16,DBLP:conf/emnlp/ChanLYCHZY19,DBLP:conf/emnlp/BakO19}. Despite users can be identified by their IDs, the personalized performance is limited because no user-related information is used in the model. Therefore, another group of researchers proposed assigning explicit profiles (personas) for chatbots to generate personalized responses. For example, 
\citet{DBLP:conf/acl/KielaWZDUS18} published the PERSONA-CHAT dataset, in which each user is assigned with several persona description sentences and the conversations between users are collected. With explicit profile descriptions, chatbots can learn to generate more consistent and informative dialogues. On this dataset, many methods have achieved encouraging performance, such as variational autoencoders~\cite{DBLP:conf/ijcai/SongZCWL19}, pre-trained language models~\cite{DBLP:journals/corr/abs-1901-08149,DBLP:conf/acl/SongWZLL20}, and multi-task modeling~\cite{DBLP:conf/aaai/SongZH020, DBLP:conf/acl/WelleckWSC19, DBLP:conf/acl/SongWZLL20}.
These explicit persona-based methods enjoy the high quality of predefined personas, but the acquisition of these persona data is expensive and even impossible when applied to real-world systems~\cite{DBLP:conf/emnlp/ChanLYCHZY19}. 

In this work, we propose using the \textit{implicit user profile} to drive the response generation. Such user profile is implicitly learned with neural models from user dialogue history rather than predefined explicitly. Due to the fact that the user dialogue history contains abundant user-specific information and is easily accessible in real-world applications, our method is more applicable in practice.

\section{Methodology}\label{sec:model}


In this section, we first provide an overview of our proposed model DHAP. The details of each component are provided later, and the model training is introduced finally.

\subsection{The Overview of DHAP}
Suppose that for a user $u$, we have their dialogue history $H$ including a series of responses issued by $u$ and the corresponding posts: $H = ((P_1,R_1),\cdots,(P_n,R_n))$, where $n$ is the number of historical post-response pairs. 
Note that the posts $(P_1,\cdots,P_n)$ here can be issued by different users, but the responses $(R_1,\cdots,R_n)$ are all issued by the same user $u$. We call them \textit{historical posts} and \textit{historical responses} respectively in the following sections.
Under the single-turn setting, given an input post $X=(x_1,\cdots,x_{L_X})$ and the user dialogue history $H$, with sequence-to-sequence modeling, our task is to generate a personalized response $Y=(y_1,\cdots,y_{L_Y})$ as:
\begin{align}
    p\left(Y|X, u \right) =  p\left(Y|X,H \right) = \prod_{t=1}^{L_Y} p\left( y_t| y_{<t}, X, H  \right),
\end{align}
where $y_t$ denotes the word generated at the $t$-th step, and $y_{<t}$ denotes the previous generated words $(y_1,\cdots,y_{t-1})$. 
It is worth noting that the dialogue history $H$ here includes dialogues between the user $u$ and several other users, thus the history is not a multi-turn dialogue between two fixed interlocutors. 

To compute the probability $p(Y|X,H)$, we design a model called DHAP, which stands for learning user profiles from Dialogue History Automatically for Personalized chatbots.
The structure of DHAP is shown in Figure~\ref{model_figure}. We briefly introduce the key components of DHAP as follows. 
The number of the modules corresponds to the mark in the figure. 
In general, DHAP considers personalized information in both the encoder and the decoder side. 

\subsubsection{Encoder}
DHAP has two different encoders, which encode the input post and user's historical responses, respectively:

\textbf{Part (1): History encoder and general user profile}.
Since abundant personalized information (\eg, background knowledge and speaking style) is often hidden in user's historical responses, DHAP firstly establishes a Transformer-based personalized language model to encode historical responses $(R_1,\cdots,R_n)$, then summarizes the general user profile $\be^G$ based on the historical responses. As $\be^G$ does not depend on the input $X$, we call it \textit{general user profile}. ``General'' does not mean it is general among all users or cannot be updated together with time. In contrast, every user has their own profile, which can be updated once they issue a new response.

\textbf{Part (2): Personalized post encoder}. DHAP also has an encoder for the input post, which is implemented by a bidirectional GRU (BiGRU). To make the post encoder aware of the user's personalized information, we use the general user profile $\be^G$ for initialization.
Consequently, the post $X=(x_1,\cdots,x_{L_X})$ is represented by hidden state sequence $\bh^P=(\bh^P_1,\cdots,\bh^P_{L_X})$.
These representations will be dynamically aggregated as a personalized post representation $\bc_t$ by an attention mechanism at the decoding step $t$.

%
\subsubsection{Decoder}
In the decoder side, DHAP incorporates the personalized information in two perspectives:

\textbf{Part (3): User history memory and dynamic post-aware user profile}. In the general user profile, all historical responses are considered in a general view. However, for a specific input post $X$, the historical responses may play different roles. Intuitively, the historical responses relevant to $X$ are more valuable than the irrelevant ones. To highlight the historical responses relevant to the input post, DHAP builds a key-value memory of historical post-response pairs $((P_1,R_1), \cdots,(P_n,R_n))$.
Given $\bc_t$ as a personalized representation of $X$, DHAP builds a dynamic post-aware user profile $\be^D_t$ by selecting and aggregating the historical responses from the user history memory. 
As $\be^D_t$ is dynamic with different input posts $X$ and summarizes related user history information, we call it \textit{dynamic post-aware user profile}. 

\textbf{Part (4): Personalized decoder}. Finally, the personalized post representation $\bc_t$, general user profile $\be^G$, and dynamic post-aware user profile $\be^D_t$ are fused together to decode the response sequentially. Inspired by CopyNet~\cite{DBLP:conf/acl/GuLLL16}, DHAP can switch between generating a word from a generic vocabulary~($p(y_t|m_g)$) and copying a word from a user's personalized vocabulary~($p(y_t|m_c)$) as:
\begin{align}
    p( y_t| y_{<t}, X, H  ) &= p(m_g)p(y_t|m_g) + p(m_c)p(y_t|m_c),\label{eq_decode_prob}
\end{align}
where $p(m_g)$ and $p(m_c)$ are computed by our designed decoding switcher. $p(y_t|m_g)$ and $p(y_t|m_c)$ are calculated based on same inputs(\eg, $\bc_t$, $\be^G$ and $\be^D_t$) with different functions.

In the remaining part of this section, we will introduce these four components of DHAP in detail.
\subsection{History Encoder and General User Profile}\label{sec:general}
Based on our observation, there is a large amount of personalized information in the user's historical responses. For example, a fan of cricket may talk a lot about cricket topics with others. 
Furthermore, different users can hold different speaking styles, such as enjoying speaking slang.
Therefore, our first idea is to devise a model for learning such personalized language information in a general view.

Inspired by the strong ability of Transformer~\cite{DBLP:conf/nips/VaswaniSPUJGKP17} to aggregate context and model sequences, we use a Transformer encoder to learn contextual representations of historical responses. 
In particular, we first add special tokens and concatenates all $n$ historical responses as $R^{w} = [\langle{\rm CLS}\rangle;R_1;\langle{\rm SEP}\rangle;\cdots;R_n;\langle{\rm SEP}\rangle]$, where each response $R_i=(r^i_1,\cdots,r^i_{L_{R_i}})$ contains $L_{R_i}$ words, and [;] is the concatenation operation. A ``$\langle{\rm SEP}\rangle$'' token is added at the tail of each response for segmentation, while a ``$\langle{\rm CLS}\rangle$'' token is added at the sequence head for summary. Then we map all words and special tokens into embeddings, and learn their contextual representations:
\begin{align}
    \left[\be^G;\bE^R\right] &= {\rm Transformer}_{N}\left( \left[\be_{{\rm CLS}};\bR_1;\be_{{\rm SEP}};\cdots;\bR_n;\be_{{\rm SEP}}\right]\right) \label{transformer_response}, \\
    \bR_i &= [\be^{r_i,{\rm fuse}}_1; \cdots; \be^{r_i,{\rm fuse}}_{L_{R_i}}], 
\end{align}
where $\be^{r_i,{\rm fuse}}_j$ is the sum of the word embedding, segment embedding, and position embedding of the $j$-th word.
These three embeddings are used in a similar way like BERT~\cite{DBLP:conf/naacl/DevlinCLT19}. 
The representation of ``$\langle {\rm CLS}\rangle$'' token ($\be^G$) summarizes the information in the entire historical responses, thus we call it the general user profile. $\bE^R$ contains contextual representations of words in historical responses. 
For $R_i$, we denote its contextual representations as $(\be^{r_i}_1,\cdots,\be^{r_i}_{L_{R_i}})$. 
${\rm Transformer}_{N}$($\cdot$) is a $N$-layer bidirectional Transformer encoder identical to the original implementation described in~\cite{DBLP:conf/nips/VaswaniSPUJGKP17}.

With the history encoder, we obtain: (1) a general user profile $\be^G$, which summarizes all historical responses and contains personalized information of the user; and (2) the contextual representations of words in the historical responses $\bE^R$. 
These representations will be used to build the dynamic post-aware user profile in Section~\ref{sec:dynamic}.

\subsection{Personalized Post Encoder}\label{sec:encoder}
In real-world applications, the posts are often very short, even ambiguous. Thus, building accurate encoding of the input post is difficult, which further leads to poor quality of the generated responses~\cite{DBLP:conf/acl/LiGBSGD16,DBLP:conf/acl/ShangLL15}. 
Fortunately, with personalized background knowledge, the chatbot can get more input information and is promising to better capture the semantic information of the post. 
Let us use an example to explain this: given a post ``The new MAC is so beautiful'', different users may have different understandings. For a programmer, ``MAC'' may refer to the Apple's laptop; but for a fashion girl, she may associate ``MAC'' with the lipstick. Therefore, the user's history can help distinguish the word ``MAC'' and provide more background knowledge for the input post, so that the post representation can be significantly enhanced. We call this encoder ``personalized post encoder'' - the encoded representations of the same post can be different for various users, because these users may have different profiles and different understandings of the same post.

Specifically, DHAP employs a BiGRU to encode the input post. Here we choose RNN-based architectures because they are better at capturing local correlations and encoding positional information than Transformers for short texts~\cite{DBLP:conf/ijcai/XiaZD19,DBLP:conf/conll/NeishiY19}. To facilitate the encoder with personalized information, DHAP uses the general user profile $\be^G$ to initialize hidden states of BiGRU. Given the post $X=(x_1,\cdots,x_{L_X})$, its representations $(\bh^P_1,\cdots,\bh^P_{L_X})$ are built as:
\begin{gather}
    \bh_i^P = {\rm BiGRU}(\bh_{i-1}^P, \bx_i),\quad \bh_0^P = {\rm ReLU}({\rm MLP}(\be^G)),
     \label{postencoder}
\end{gather}
where $\bx_i$ is the embedding obtained by the embedding table. 

These hidden states draw personalized information from the general user profile, and in the decoding phase, they are aggregated by the decoding hidden state $\bh_t^R$ through an attention mechanism:
\begin{gather}
    \bc_t = \sum_{i=1}^{L_X}\alpha_{t,i}\bh_i^{p}\label{attention}, \\
    \alpha_{t,i} = \frac{{\exp}\left(s_{t,i}^{rp}\right)}{\sum_{j=1}^{L_X} {\exp}(s_{t,j}^{rp})},\quad s_{t,i}^{rp} = \bv^{\top}_{rp} {\tanh}\left({\rm MLP}\left(\left[\bh_t^R;\bh_i^P\right]\right)\right). \label{atten}
\end{gather}
The detailed calculation and updating scheme of the decoding state $\bh_t^R$ at time step $t$ will be described in Section~\ref{decoder_state}.
Based on the attention mechanism, the personalized post representation $\bc_t$ can dynamically focus on some important words of the post according to the current decoding state. 
In the next section, DHAP will use $\bc_t$ to build the dynamic post-aware user profile. 

\subsection{User History Memory and Dynamic Post-aware User Profile}\label{sec:dynamic}
With the general user profile, DHAP can capture personalized information of a user in a general view. However, when faced with a new input post, the historical data may play different roles. 
For example, when a crazy fan of cricket meets a post about cricket, they will be talkative and post a lot of things. But when they face other daily topics, they may behave much more gently. 
Hence, it is valuable to dynamically select the information that is most relevant to the input, and the chatbot can behave differently as the input varies.
Following this idea, we propose to dynamically aggregate historical responses that are highly relevant to the current input post, and leverage them as a reference to drive the response generation.

To achieve this, DHAP uses a key-value memory network~\cite{DBLP:conf/emnlp/MillerFDKBW16} to store the user's historical post-response pairs. 
Then the personalized post representation is used as the query to select highly related keys (historical posts) from the user history memory, and their corresponding values (historical responses) are aggregated as the dynamic post-aware user profile. 

\subsubsection{User History Memory}
We firstly transform the historical post-response pairs into key-value pairs and build the memory.

As we discussed earlier, the historical posts are usually issued by different users. Thus the language style and topic of them may be various. Under this circumstance, it is more reasonable to treat them independently, so DHAP applies a BiGRU to represent each historical post, respectively. In our implementation, this BiGRU shares parameters with the personalized post encoder (introduced in Section~\ref{sec:encoder}). Consider the $i$-th historical post $P_i$, its representation is computed by a summing pooling over the word dimension as $\bp_i^c = \sum_{j=1}^{L_{P_i}} \bh^{i}_j$, where $\bh^{i}_j$ is the hidden state of the BiGRU for the $j$-th word in $P_i$.
For all historical posts $(P_1,\cdots,P_n)$, their representations are denoted as $(\bp_1^c,\cdots,\bp_n^c)$. Similarly, the representation of historical responses is also computed by the same pooling strategy as $\br_i^c = \sum_{j=1}^{L_{R_i}} \be^{r_i}_j$, where $\be^{r_i}_j$ is the contextual representation of the $j$-th word in $R_i$. Different from historical posts, the contextual word representations of historical responses are obtained by the history encoder in Equation (\ref{transformer_response}). Thus, all historical responses are represented as $(\br_1^c, \cdots,\br_n^c)$.

Finally, we build the user history memory by using the post representations as key and the corresponding response representations as value, \ie, $\left\{\bk_1^{M}:\bv_1^{M},\cdots,\bk_n^{M}:\bv_n^{M} \right\} = \left\{\bp_1^c:\br_1^c,\cdots,\bp_n^{c}:\br_n^{c} \right\}$.

\subsubsection{Dynamic Post-aware User Profile}
After building the user history memory, DHAP can select and aggregate the most relevant historical responses and build the dynamic post-aware user profile based on the input post. 
Specifically, the personalized representation of the input post obtained in Equation~(\ref{attention}) is used as the query and attends to the memory keys to find the most relevant historical posts. The relevance is measured by the attention weights. Then the corresponding historical responses are summed up based on the normalized weights to construct the dynamic profile $\be^D_t$:
\begin{gather}
    \be^D_t = \sum_{i=1}^{n} \beta_{t,i}\bv^{M}_i, 
\end{gather}
where $\beta_{t,i}$ is the attention weight of the $i$-th historical response based on the personalized post representation $\bc_t$ in a similar way like Equation~(\ref{atten}).
Note that the dynamic post-aware user profile is computed at each decoding time step $t$. 
The most relevant information hidden in historical responses can thus be selected to help response generation. 

\subsection{Personalized Decoder}\label{sec:decoder}
Finally, the generation probability of responses can be calculated by the personalized post representation $\bc_t$, the general user profile $\be^G$, and the dynamic post-aware user profile $\be^D_t$. Inspired by CopyNet~\cite{DBLP:conf/acl/GuLLL16}, in addition to leveraging the personalized information captured by the implicit user profile, we construct a personalized vocabulary so that the model is allowed to directly select personalized words that the user frequently used in history. 

Specifically, the probability of the word $y_t$ generated in the personalized decoder is computed as Equation~(\ref{eq_decode_prob}), where $p\left(m_g\right)$ is the probability of general decoding mode and $p\left(m_c\right)$ is the probability of copy decoding mode. They are computed by our proposed decoding switcher. $p\left(y_t|m_g\right)$ and $p\left(y_t|m_c\right)$ are the probabilities of generating $y_t$ under two modes, respectively.

It is worth noting that: (1) The switching probability $p\left(m_g\right)$ and $p\left(m_c\right)$ are both in $[0,1]$. Thus, it is a ``soft'' decoding switcher. (2) The generic vocabulary also contains words in the personalized vocabulary. The generation probability of $y_t$ is obtained by the sum of probabilities under two decoding modes. Therefore, DHAP is just biased to the personalized words rather than lost in it.

\subsubsection{Decoding switcher}\label{decoding_switcher}
The decoding switcher determines the probability of the two decoding modes, \ie, predicting a word from the generic vocabulary to maintain sentence fluency, or copying a word directly from the personalized vocabulary to make the response more informative and personalized. Specifically, DHAP computes the switching probability based on the matching degree between the decoder state and the concatenation of the personalized post representation, general user profile, and dynamic post-aware user profile, and further calculates two decoding mode probabilities.
\begin{align}
    [p(m_g),p(m_c)] &= {{\rm Softmax}}(\bd_t), \\
    \bd_t &= {\rm MLP}([\bh_t^R;\bc_t;\be^G;\be^D_t]),
\end{align}
where $\bh_t^R$ is the decoding hidden state at step $t$, and $\bd_t \in \mathbb{R}^{2}$ is the matching degree vector to estimate the two mode probabilities. The softmax function guarantees that $p(m_g)+p(m_c)=1$.

\subsubsection{Personalized general decoding}\label{general_decoding}
While general decoding, the decoder should predict a word $y_t$ from the generic vocabulary:
\begin{align}
    p(y_t|m_g) &= {\rm Softmax}({\rm MLP}([ \bh_t^R;\bc_t;\be^G;\be^D_t ])).
\end{align}
\subsubsection{Personalized copy decoding}\label{copy_decoding}
The personalized vocabulary of user $u$ is composed of the words that appear in their historical responses. DHAP can directly select a word from this vocabulary to generate a more personalized response. Inspired by copy mechanism~\cite{DBLP:conf/acl/GuLLL16}, the probability of selecting a word $y_t$ is computed as: 
\begin{align}
    p\left(y_t|m_c\right) = \sum_{i:r_i=y_t} \gamma_{t,i},
\end{align}
where $\gamma_{t,i}$ is the attention weight calculated by the personalized post representation $\bc_t$ attentively reading the representation of historical responses $\bE^R$ with the same attention process in Equation~(\ref{attention}).

\subsubsection{Decoder state updating} \label{decoder_state}
DHAP applies a GRU as the decoder. The hidden state at decoding step $t$ is calculated as:
\begin{align}
    \bh_t^R = {{\rm GRU}}(\bh_{t-1}^R, [\by_{t-1};\bc_t;\be^G;\be_t^D]),
\end{align}
where $\by_{t-1}$ is the embedding vector of the last generated word. 
The decoding states are initialized by the last hidden state of the personalized post encoder:
\begin{align}
    \bh_0^R = {{\rm ReLU}}({\rm MLP}(\bh^P_{L_X})).
\end{align}

\subsection{Training and Optimization}\label{section_training}
Our goal is to maximize the generation probability of the target response given the input post and user's dialogue history.  
A length penalty is applied as~\cite{DBLP:conf/acl/LiGBSGD16} to alleviate the generation of meaningless responses. As a result, the loss function of DHAP is defined as:
\begin{align}
    \mathcal{L} = - \sum_{t=1}^{L_Y} {\rm log}\left[p\left(y_t| y_{<t}, X, H\right)\right] - \eta L_Y,\label{loss}
\end{align}
where $\eta$ is a hyper-parameter to control the associated length penalty
weight. $p\left(y_t| y_{\textless t}, X, H\right)$ is the generation probability of word $y_t$ based on the given input post and user's history, which is computed in Equation~(\ref{eq_decode_prob}). All parameters are optimized by the loss function and the whole model is trained in an end-to-end manner.


\section{Experiments}\label{sec:exp}
\subsection{Datasets}
Although there are many public datasets for response generation, as far as we know, none of them contain user identification.
To collect each user's dialogue history and evaluate the effectiveness of our model, we use two datasets extracted from two open online chatting forums, \ie, Weibo and Reddit. The two datasets contain massive dialogue utterances (\ie, post-response pairs) and user identification information, thus we can sample the data by users. To guarantee enough personalized information, we retrieve users with more than ten utterances to maintain an effective dialogue history. Each utterance is used as the target response for generation, while its former responses and the corresponding posts are treated as the dialogue history.  We divide the utterances by users into 8:1:1 as training, validation, and test set respectively in time order. 
Besides, given a user, we ensure that the time of its records in the validation set and test set are behind the records in the training set. 

\textbf{Weibo dataset} is a subset of PChatbotW~\cite{pchatbot_sigir}, which is collected from Weibo for the one-year period beginning from Sept. 10, 2018. 
On Weibo, a user can post short messages visible to the public, which will be referred to as posts. Other users can make comments on a published post, which will be referred to as response. 
For data cleaning, we remove hashtags, URLs, emoticons, and duplicate text as~\cite{pchatbot_sigir}. We also remove the utterances whose length is less than five words or more than 100 words. We use comparable scales of samples with~\cite{DBLP:conf/emnlp/ChanLYCHZY19} to conduct our experiments. It comprises 300,000 users and 8,618,374 utterances, in total 31M words.  

\textbf{Reddit dataset} is extracted from comment chains scraped from Reddit from Dec. 1, 2015 to Oct. 30, 2018~\cite{DBLP:conf/acl/ZhangSGCBGGLD20}. Since the Reddit discussions can be naturally expanded as tree-structured reply chains, we pair the parent node with all its child nodes respectively, and construct multiple post-response pairs.
we treat the parent node and the child node as the post and response, respectively.
As a result, a parent node can be a submission or a comment, while a child node only refers to a comment. For each submission, we use its title as the post text.
We clean the raw data by removing instances containing word repetitions, offensive words, or multi-language sentences.
It contains 315,340 users and 24,162,464 utterances, in total 55M words. 

\subsection{Baselines}
We evaluate the performance of our approach by comparing it with four groups of highly related and strong baseline methods: 

(1) \textit{Non-personalized response generation models}. \textbf{Seq2SeqWA}~\cite{DBLP:journals/corr/BahdanauCB14} is a standard GRU-based Seq2Seq model with attention mechanism. 
\textbf{MMI}~\cite{DBLP:conf/naacl/LiGBGD16} is a Seq2SeqWA using Maximum Mutual Information as loss function to improve diversity.

(2) \textit{Personalized models using user ID embeddings}. \textbf{Speaker}~\cite{DBLP:conf/acl/LiGBSGD16} is also based on Seq2SeqWA but using user ID embeddings as additional input to the decoder. 
\textbf{PersonaWAE}~\cite{DBLP:conf/emnlp/ChanLYCHZY19} is built on an augmented Wasserstein autoencoder. It utilizes user ID embeddings for building a personalization Gaussian mixture distribution, and fuses personalization in the decoder.

(3) \textit{Personalized models using explicit user profiles}. Since no explicit user profiles are given in our datasets, we use the historical responses of users as their persona texts. \textbf{GPMN}~\cite{DBLP:conf/acl/KielaWZDUS18} enhances the Seq2SeqWA with a memory module, which encodes each piece of persona description as an individual memory representation. It uses the input message as the query to aggregate and incorporate the memory representations for response generation. 
\textbf{PerCVAE}~\cite{DBLP:conf/ijcai/SongZCWL19} uses the user profile descriptions as conditions and applies a conditional variational autoencoder to generate diverse responses.

(4) \textit{Personalized models using implicit user profiles}.
Since no existing methods consider mining user profiles from dialogue history implicitly, we adapt several state-of-the-art multi-turn response generation models to personalized response generation. 
We replace the dialogue context in the original models by the user’s
historical post-response pairs.
\textbf{VHRED-P}~\cite{DBLP:conf/aaai/SerbanSLCPCB17} extends the hierarchical recurrent encoder-decoder with a latent variable to model the complex dependencies among multiple utterances in the context. 
\textbf{ReCoSa-P}~\cite{DBLP:conf/acl/ZhangLPGC19} applies a self-attention mechanism to measure the relevance between the response and each utterance in the context. 
\begin{table*}[t!]
    \centering
    \small
    \caption{Automatic evaluation results of all models. All models are categorized into four groups: (1) non-personalized; (2) using user ID; (3) using explicit user profile; and (4) using dialogue history.
    ``$\dagger$'' denotes the result is significantly worse than our method DHAP in t-test with $p \textless 0.05$ level. 
    The best results are in  \textbf{bold} and the second best results are \underline{underlined}.}
    \label{table_a}
    \begin{tabular}{llllllllllll}
    \toprule
        \multirow{2}{*}{Dataset} & \multirow{2}{*}{Model}  & \multicolumn{3}{c}{Word Overlap} & \multicolumn{2}{c}{Diversity} & \multicolumn{3}{c}{Embedding Similarity} & \multicolumn{2}{c}{Personalization}  \\
    \cmidrule(lr){3-5} \cmidrule(lr){6-7}  \cmidrule(lr){8-10} \cmidrule(lr){11-12}
        & & BLEU-1 & BLEU-2 & ROUGE-L & Dist-1 & Dist-2 & Average & Extrema & Greedy  & P-F1(\%) & P-Cover \\
    \midrule
    \multirow{9}{*}{Weibo} & (1) Seq2SeqWA & $3.335^\dagger$ & $0.294^\dagger$ & $8.740^\dagger$ & $0.935^\dagger$ & $2.184^\dagger$ & $0.321^\dagger$ & $0.266^\dagger$ & $0.254^\dagger$ &  $1.698^\dagger  $&  $0.041^\dagger  $ \\
    & (1) MMI & $3.632^\dagger$ & $0.095^\dagger$ & $5.317^\dagger$  & $10.714^\dagger$ & $43.479^\dagger$ & $0.477^\dagger$ & $0.695^\dagger$ & $0.305$ &  $1.874^\dagger$ & $0.054^\dagger$ \\
    & (2) Speaker & $4.997^\dagger$ & $0.113^\dagger$ & $7.993^\dagger$  & $6.035^\dagger$ & $19.017^\dagger$ & $0.492^\dagger$ & $0.712^\dagger$ & $0.311$ &  $2.119^\dagger$ & $0.082^\dagger$  \\
    & (2) PersonaWAE & $3.503^\dagger$ & $0.155^\dagger$ & $11.305^\dagger$  & $2.493^\dagger$ & $19.716^\dagger$ & $\underline{0.513}^\dagger$ & $\underline{0.724}^\dagger$ & $0.307$ &  $5.108^\dagger$ & $\underline{0.093}^\dagger$ \\
    & (3) GPMN & $4.901^\dagger$ & $0.696^\dagger$ & $8.090^\dagger$  & $11.726^\dagger$ & $32.734^\dagger$ & $0.353^\dagger$ & $0.391^\dagger$ & $0.301^\dagger$ &  $4.512^\dagger$&  $0.084^\dagger$  \\
    & (3) PerCVAE & $5.115^\dagger$ & $0.299^\dagger$ & $7.952^\dagger$  & $\underline{14.095}^\dagger$ & $\underline{49.739}^\dagger$ & $0.469^\dagger$ & $0.659^\dagger$ & $0.299^\dagger$  & $3.817^\dagger$ & $0.086^\dagger$   \\
    & (4) VHRED-P & $6.992^\dagger$ & $0.709^\dagger$ & $10.695^\dagger$  & $2.122^\dagger$ & $7.874^\dagger$ & $0.437^\dagger$ & $0.560^\dagger$ & $0.307$  & $5.459^\dagger$ & $0.065^\dagger$  \\
    & (4) ReCoSa-P & $\underline{7.266}^\dagger$ & $\underline{0.844}^\dagger$ & $\underline{11.469}^\dagger$  & $1.271^\dagger$ & $4.442^\dagger$ & $0.419^\dagger$ & $0.510^\dagger$ & $\underline{0.312}$ & $\underline{5.717}^\dagger$ & $0.061^\dagger$ \\
    & (4) DHAP (ours) & $\bm{9.324}$ & $\bm{0.894}$ & $\bm{14.122}$ & $\bm{15.175}$ & $\bm{58.806}$ & $\bm{0.523}$ &$\bm{0.747}$ & $\bm{0.313}$  & $\bm{7.013}$ & $\bm{0.144}$  \\
     \midrule
    \multirow{9}{*}{Reddit} &  {(1) Seq2SeqWA} & $1.819^\dagger$ & $0.023^\dagger$ & $4.069^\dagger$ & $5.203^\dagger$ & $19.485^\dagger$ & $0.545^\dagger$ & $0.554^\dagger$ & $0.472^\dagger$  & $0.516^\dagger$ & $0.029^\dagger$ \\
    & (1) MMI & $2.065^\dagger$ & $0.011^\dagger$ & $3.784^\dagger$  & $5.914^\dagger$ & $31.093^\dagger$ & $0.543^\dagger$ & $0.607^\dagger$ & $0.454^\dagger$  &  $0.828^\dagger$ & $0.038^\dagger$ \\
    & (2) Speaker & $2.642^\dagger$ & $0.054^\dagger$ & $4.469^\dagger$  & $8.951^\dagger$ & $34.187^\dagger$ & $0.538^\dagger$ & $0.606^\dagger$ & $0.457^\dagger$  &  $1.455^\dagger$ & $0.031^\dagger$ \\ 
    & (2) PersonaWAE & $2.637^\dagger$ & $0.113^\dagger$ & $8.199^\dagger$  & $1.758^\dagger$ & $25.915^\dagger$ & $0.629^\dagger$ & $\underline{0.685}^\dagger$ & $0.442^\dagger$  &  $3.392^\dagger$ & $0.032^\dagger$ \\
    & (3) GPMN &  $2.686^\dagger$ & $0.376^\dagger$ & $4.776^\dagger$  & $\underline{12.325}^\dagger$ & $35.762^\dagger$ & $0.406^\dagger$ & $0.331^\dagger$ & $0.358^\dagger$  & $3.026^\dagger$ & $0.037^\dagger$   \\
    & (3) PerCVAE & $5.933^\dagger$ & $0.576^\dagger$ & $8.112^\dagger$  & $9.631^\dagger$ & $\underline{40.213}^\dagger$ & $\underline{0.637}^\dagger$ & $0.649^\dagger$ & $0.499^\dagger$  &  $3.456^\dagger$ & $0.040^\dagger$ \\
    & (4) VHRED-P & $5.802^\dagger$ & $0.648^\dagger$ & $8.345^\dagger$  & $2.750^\dagger$  & $30.756^\dagger$ & $0.558^\dagger$ & $0.635^\dagger$ & $0.472^\dagger$  & $3.772^\dagger$ & $\underline{0.047}^\dagger$ \\ 
    & (4) ReCoSa-P & $\underline{6.113}^\dagger$ & $\underline{0.686}^\dagger$  & $\underline{8.899}^\dagger$ &$2.593^\dagger$ & $25.767^\dagger$ & $0.574^\dagger$ & $0.632^\dagger$  & $\underline{0.510}^\dagger$  & $\underline{3.998}^\dagger$ & $0.044^\dagger$ \\
    & (4) DHAP (ours) & $\bm{6.858}$ & $\bm{0.737}$ & $\bm{11.720}$  & $\bm{18.707}$ & $\bm{66.932}$ & $\bm{0.709}$ &$\bm{0.721}$ & $\bm{0.539}$  &  $\bm{4.639}$ &  $\bm{0.111}$ \\
    \bottomrule
    \end{tabular}
\end{table*}

\begin{table}[t!]
    \centering
    \small
    \caption{Human evaluation results on Weibo dataset.  
    ``$\dagger$'' denotes the result is significantly worse than our method in t-test with $p \textless 0.05$ level. 
    The best results are in  \textbf{bold} and the second best results are \underline{underlined}. The Fleiss Kappa is 0.42.}
    \label{table_human}
    \begin{tabular}{lccc}
    \toprule
        Model & Readability & Informativeness & Personalization \\
    \midrule
        (1) Seq2SeqWA & $2.10^\dagger$ & $1.85^\dagger$ & $0.19^\dagger$\\
        (1) MMI      &$2.06^\dagger$ & $1.88^\dagger$ & $0.23^\dagger$\\ 
        (2) Speaker  & $\underline{2.14}^\dagger$& $1.93^\dagger$ & $0.25^\dagger$ \\
        (2) PersonaWAE & $2.07^\dagger$ & $1.99^\dagger$ & $0.36^\dagger$ \\
        (3) GPMN     &$2.12^\dagger$  &$1.92^\dagger$ &$0.35^\dagger$\\
        (3) PerCVAE  &$2.04^\dagger$  & $\underline{2.01}^\dagger$ &$0.39^\dagger$\\
        (4) VHRED-P  &$2.09^\dagger$  &$1.96^\dagger$ &$\underline{0.47}^\dagger$\\
        (4) ReCoSa-P &$2.12^\dagger$  &$1.93^\dagger$ &$0.44^\dagger$\\
        (4) DHAP (ours)    &$\bm{2.26}$  &$\bm{2.09}$ & $\bm{0.56}$\\
        \midrule
        Ground-truth &2.69  &2.35 &0.84\\
    \bottomrule
    \end{tabular}
\end{table}
\subsection{Evaluation Metrics}

\textbf{Automatic Evaluation}: We consider several automatic metrics in different perspectives to jointly evaluate the generated responses.
(1) We use BLEU-1, BLEU-2~\cite{DBLP:conf/acl/PapineniRWZ02}, and ROUGE-L~\cite{DBLP:conf/acl/LinO04} to measure word overlaps between the generated response and ground truth. A higher value of these metrics indicates a higher word-level similarity between the generated response and the golden response.
(2) Following~\cite{DBLP:conf/naacl/LiGBGD16}, we employ Dist-1 and Dist-2 to evaluate the diversity of the generated response. Responses with more distinct unigrams/bigrams will have higher Dist-1/Dist-2. 
(3) As suggested by~\cite{DBLP:conf/emnlp/ChanLYCHZY19}, we use three embedding-based metrics to measure the semantic relevance between the generated response and the ground-truth response. Concretely, we use the bag-of-words embeddings to represent both the generated and ground-truth response, and calculate their average similarity (Ave.), greedy similarity (Gre.), and extrema similarity (Ext.). The pre-trained word embeddings for Weibo and Reddit corpus are offered by \citet{P18-2023} and \citet{DBLP:conf/emnlp/PenningtonSM14}, respectively.
(4) Furthermore, since the goal of our model is to leverage user history for personalization, we evaluate the personalized performance by measuring how much information in the dialogue history is reflected in the generated response. 
Following~\cite{DBLP:conf/ijcai/LianXWPW19,DBLP:conf/dasfaa/LvFWZY20}, we use Persona F1 (P-F1) to measure the unigram F1 between the generated response and user's historical responses. Thus, the more historical words the generated response contains, the higher P-F1 we will get.
Since the importance of the shared words can be different, following~\cite{DBLP:conf/ijcai/SongZCWL19}, we further use Persona Coverage (P-Cover) to measure the IDF-weighted word overlap between generated response and dialogue history. 
Specifically, for $n$ historical responses $\left\{R_1,\cdots,R_n\right\}$ and the generated response $Y$, P-Cover is defined as:
\begin{align}
  {\rm P\text{-}Cover} = {{\rm max}}_{j \in \left[1,n\right] } \frac{ \sum_{w_k \in W_j}{\rm IDF}(w_k)}{|Y|},
\end{align}
where $W_j$ is the set of shared words between $R_j$ and $Y$. 


\begin{table*}[t!]
    \centering
    \small
    \caption{Performance of ablation models on Weibo dataset. ``$\dagger$'' denotes the result is significantly worse than our method in t-test with $p \textless 0.05$ level.  The best results are denoted in \textbf{bold} font.}
    \label{table_ablation}
    \begin{tabular}{p{0.1\textwidth}llllllllll}
    \toprule
         \multirow{2}{*}{Model}  & \multicolumn{3}{c}{Word Overlap} & \multicolumn{2}{c}{Diversity} & \multicolumn{3}{c}{Embedding Similarity}  & \multicolumn{2}{c}{Personalization} \\
    \cmidrule(lr){2-4} \cmidrule(lr){5-6}  \cmidrule(lr){7-9} \cmidrule(lr){10-11}
         & BLEU-1 & BLEU-2 & ROUGE-L & Dist-1 & Dist-2 & Average & Extrema & Greedy  & P-F1(\%) & P-Cover \\
    \midrule
     DHAP & $\bm{9.324}$ & $\bm{0.894}$ & $\bm{14.122}$ & $\bm{15.175}$ & $\bm{58.806}$ & $\bm{0.523}$ &$\bm{0.747}$ & $\bm{0.313}$   & 7.013 & 0.144 \\
     \midrule
     \quad \textit{w/o} G & $7.726^\dagger$ & $0.801^\dagger$ & $11.815^\dagger$ & $12.176^\dagger$ & $49.808^\dagger$ &  $0.495^\dagger$ &  $0.707^\dagger$ & $0.294^\dagger$  & $6.179^\dagger$ & $0.107^\dagger$   \\
     \quad \textit{w/o} D & $8.503^\dagger$ & $0.855^\dagger$  & $12.610^\dagger$ & $13.699^\dagger$ & $54.623^\dagger$ & $0.499^\dagger$ & $0.713^\dagger$ & $0.303^\dagger$  &   $6.286^\dagger$ & $0.109^\dagger$   \\
     \quad \textit{w/o} PC & 8.830 & $0.868^\dagger$ & 13.981 & 14.457 & $56.263^\dagger$ & $0.503^\dagger$ & $0.728^\dagger$ & $0.301^\dagger$  & 6.884   & $0.120^\dagger$\\
          \midrule
    \quad \textit{w/o} GEN & $4.982^\dagger$ & $0.328^\dagger$  & $9.571^\dagger$ & $9.051^\dagger$ & $32.566^\dagger$ & $0.478\dagger$ & $0.571^\dagger$ & $0.276^\dagger$ & \textbf{9.331} & $\textbf{0.165}$  \\
    \quad \textit{w/o} COP & $8.347^\dagger$ &  $0.837^\dagger$ & $12.585^\dagger$ & $13.487^\dagger$ & $52.087^\dagger$ & $0.499^\dagger$ &$0.717^\dagger$   & $0.298^\dagger$  & $6.234^\dagger$ & $0.110^\dagger$  \\
    \quad \textit{w} FIX & $8.549^\dagger$ & $0.855^\dagger$ & $12.871^\dagger$ & $13.904^\dagger$ & $54.539^\dagger$ & $0.496^\dagger$ & $0.716^\dagger$  & $0.301^\dagger$  & $6.326^\dagger$ & $0.113^\dagger$  \\  
    \bottomrule
    \end{tabular}
\end{table*}

\textbf{Human Evaluation}:
The automatic evaluation metrics can measure the quality of the generated response with respect to the ground-truth. However, due to the diversity of human dialogues, a response different from the ground-truth may also be acceptable. Thus, we randomly sample 100 test samples to conduct human evaluations. We present the generated responses, the corresponding post, and the user's historical post-response pairs to three well-educated annotators. The annotators will evaluate the quality of the generated responses in a double-blind fashion. Following~\cite{DBLP:conf/emnlp/ChanLYCHZY19}, the evaluation criterion includes: (1) Readability, which measures the grammatical correctness and smoothness of generated responses; (2) Informativeness, which measures whether the responses are informative or trivial; and (3) Personalization, which measures if the response can reflect personalized information (sharing some information with the history of the user). For the former two perspectives, we use a score 1/2/3 for bad/normal/good quality. For personalization, we use the score 0/1 to judge whether a response reflects personalized information or not. 
The Fleiss Kappa is 0.42 that indicates the annotators achieve a substantial agreement.

\subsection{Implement Details}
To determine the parameters of the model, we conducted multiple sets of experiments. The final parameters are selected as follows. For all datasets, we use 512 as the hidden size of GRU, 0.001 as the learning rate. The hidden size and number of heads of Transformer are 256 and 8. The number of Transformer layers $N=6$. The history length is set to 25. The vocabulary size is limited to 40,000. The word embedding dimension is 300/100 for Weibo/Reddit datasets, respectively. We use the Adam optimizer with a batch size of 256. We train all models for 10 epochs and select the best model based on the validation results on BLEU-1.

\subsection{Experimental Results}
\subsubsection{Automatic Evaluation}
All evaluation results under automatic metrics are reported in Table~\ref{table_a}. We can observe that: 

(1) Among all models, \textbf{DHAP achieves the best results in terms of all evaluation metrics.} DHAP improves performance with a large margin over two strongest baselines VHRED-P and ReCoSa-P, which can also learn implicit user profile. 
Concretely, DHAP significantly outperforms ReCoSa-P by $28.4\%/12.2\%$ improvements in BLEU-1 on Weibo/Reddit dataset. The reason for the improvement reduction on Reddit set is that it has a larger scale and more varied conversations, which leads to more noise. Besides, for the embedding similarity metrics, DHAP also outperforms the best baselines. These results demonstrate that DHAP can generate more semantically relevant responses to the ground-truth by leveraging user's history. Furthermore, DHAP has dramatic improvements of Dist-1/2, indicating DHAP can generate more informative and diverse responses based on the personalized information. All these results prove that learning implicit user profiles from user's dialogue history can improve the quality of generated responses.

(2) All personalized methods outperform non-personalized methods, indicating that \textbf{personalization is helpful for generating more informative and relevant responses}. Seq2SeqWA generally has the lowest performance, reflecting that the semantic information in the post is insufficient for generating an informative response. MMI improves the diversity performance significantly, but loses some ability on modeling semantic relevance, as it changes the training objective. Speakers and PersonaWAE use user ID embeddings to identify different users for personalization, and outperform the non-personalized methods. The explicit persona-based model GPMN and PerCVAE show comparable performance to the user ID embedding based baselines. A potential reason is that they are designed for leveraging explicit user profile, which is usually of high quality. In our case, they are only provided with the user's historical responses, which are much noisy. Therefore, existing personalized methods for explicit user profile is not appropriate for dealing with the implicit user profile contained in the user history.

(3) Among all personalized methods, \textbf{the ones using implicit user profile perform better.} VHRED-P and ReCoSa-P show better performance on most metrics, confirming that the dialogue history can be used to mine implicit user profile for a specific user. However, these two methods are originally proposed for multi-turn dialogue generation. Their performance in personalized tasks is limited because the dialogue history covers far more aspects than the context in multi-turn dialogue. On the contrary, our DHAP models the personalized information in both encoder and decoder side, and consider the implicit user profile in both general and dynamic style.
Hence, DHAP can achieve significant improvements compared with existing personalized baselines.

\subsubsection{Human Evaluation}
We also conduct a human evaluation for all models on Weibo dataset. The results are shown in Table~\ref{table_human}. Generally, DHAP achieves significant improvements in terms of all perspectives, which is consistent with the results on automatic metrics. In particular, we find that DHAP is much better than ReCoSa-P in terms of personalization. This is because DHAP learns the implicit user profile more comprehensively and enhances the influence of personalized words directly in the decoder. DHAP also performs better than other baselines in terms of readability, which shows that DHAP is better at language understanding with the help of user history. 
Besides, the informativeness of responses generated by DHAP is also improved. This demonstrates that leveraging personalized information is effective to generate more meaningful responses. 

In summary, the automatic and human evaluation results strongly verify that \textbf{dialogue history is suitable to build the user profile implicitly, and leveraging implicit user profiles is effective to generate meaningful and personalized responses, further achieving a personalized chatbot}.

\begin{figure}
\centering
\includegraphics[scale=0.4]{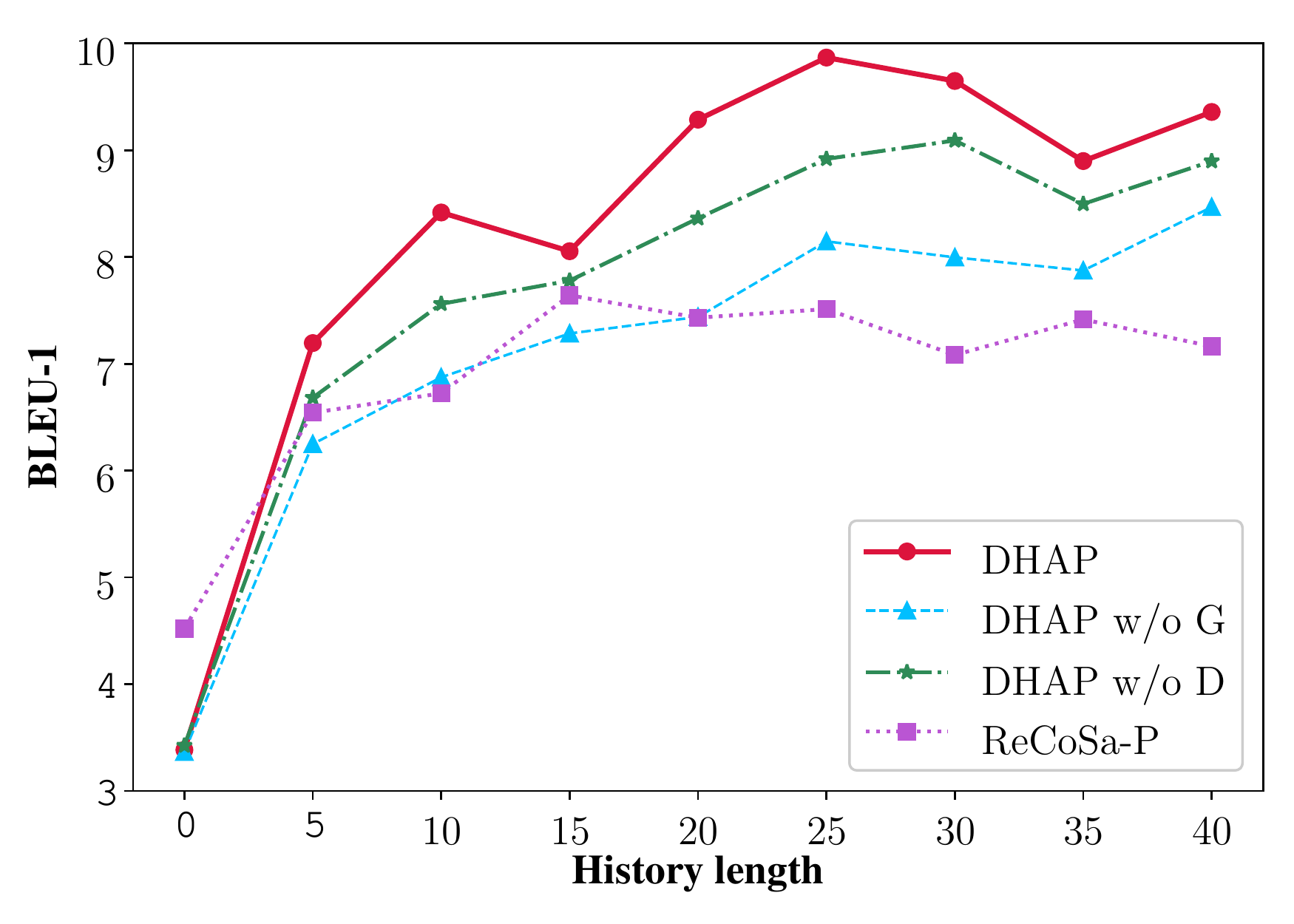}
\caption{Effectiveness of DHAP on users with different lengths of history for Weibo dataset. }
\label{fig:figure_length}
\end{figure}

\subsection{Further Analysis}
We further analyze the influence of different modules (Section~\ref{sec:ab}) and the performance over different history lengths (Section~\ref{sec:length}). Both of these experiments are performed on Weibo dataset.
\subsubsection{Ablation Study}\label{sec:ab}
DHAP learns several personalized user profiles based on the dialogue history and designs a decoding switcher and two decoding strategies in the personalized decoder. We remove one of them once a time to analyze its contribution. The experiment results on Weibo dataset are shown in Table~\ref{table_ablation}.

\textbf{The ablation on personalized representations.}  Three settings are considered: (1) \textbf{without G}: the general user profile $\be^G$ is not used; (2) \textbf{without D}: the dynamic post-aware user profile $\be^D_t$ is not used; and (3) \textbf{without PC}: the post encoder is non-personalized, namely initializing the post encoder with random states rather than the general user profile. 

The results show that all of the personalized representations are useful. Specifically, removing the general user profile causes the most decline in all metrics, which confirms the necessity and contribution of it on summarizing personalized information in a general view. The performance degradation caused by removing the dynamic post-aware user profile shows that selecting historical responses relevant to the input post contributes to the further enhancement of user modeling. The influence of removing personalization in post encoder is relatively smaller. It proves that using the user profile to enhance the understanding of the current post is effective but limited, since such information is only provided at the beginning step and decreases with the hidden state update.  

\textbf{The ablation of components in the personalized decoder.} 
We test the following variants of our model:  (1) \textbf{without GEN}: the general decoding is banned; (2) \textbf{without COP}: the copying mode is banned; (3) \textbf{with FIX}: the probability of two modes are fixed. 
Specifically, the general decoding probability is set as 0.8 and the copy decoding probability is 0.2, these probabilities are set according to the best results of DHAP in our preliminary experiments.

It can be seen that the results of three variants all underperform the whole framework.
Without general decoding, the performance of DHAP drops sharply in terms of all metrics except personalization metrics. Specifically, it drops 46.55\% in terms of BLEU-1.
This indicates that only copying words from personalized vocabulary is unable to generate a suitable response, because there are lots of noises irrelevant to the current post and some general words may not be contained in the vocabulary. 
The reason for its improvement on personalized metrics is that all of the generated words are copied from the history, regardless of the significant hurt on relevance and diversity.
Thus, general decoding considering both the post, decoding states, and personalized information is necessary. However, only using general decoding also hurts the performance, which indicates the words reflecting user personalized information is also very valuable. To combine the two decoding strategies, DHAP calculates the possibilities of two decoding strategies dynamically. 
It works well in DHAP yet using fixed probabilities has lower performance.  

\subsubsection{Performance across Various Lengths of Dialogue History}\label{sec:length}
As we leverage user's dialogue history for personalization, the length of history may affect the model's performance.
To investigate the influence of history length, we test the performance of DHAP by using different numbers of historical post-response pairs. The results of BLEU-1 on Weibo set are illustrated in Figure~\ref{fig:figure_length}. We find:

(1) In general, DHAP performs better when a user has a longer dialogue history. This is consistent with our speculation as a longer dialogue history can provide more personalized information. DHAP achieves the best performance with the history length around 25. 
Unfortunately, when more than 30 historical pairs are used, the performance of DHAP becomes unstable. 
The potential reason is that more historical data may bring more noise and increase the difficulty of building an effective user profile. 
(2) 
When the history is less than 5, ReCoSa-P performs better than DHAP without general user profile. This is because the persona information is extremely limited. Under this circumstance, the more complex architecture of ReCoSa shows its superiority. 
Nevertheless, our DHAP still performs best, showing its scalability for various history lengths.


\section{Conclusion}\label{sec:conclusion}

In this work, we implemented response generation of personalized chatbots in an alternative way. 
Different from existing personalized methods, we propose the personalized model DHAP, which learns the implicit user profile automatically from large-scale user dialogue history. 
We design a personalized language model to capture the user's general interest from their historical responses and summarize the general user profile. 
To further highlight the historical responses which are relevant and valuable to the current input post, we build a history memory and construct the dynamic post-aware user profile. 
We build a personalized decoder to coordinate two personalized decoding strategies.
Experimental results confirm the effectiveness of our model on generating informativeness and personalized responses.


\begin{acks}
Zhicheng Dou is the corresponding author. This work was supported by National Natural Science Foundation of China No. 61872370 and No. 61832017, and Beijing Outstanding Young Scientist Program NO. BJJWZYJH012019100020098, and Shandong Provincial Natural Science Foundation under Grant ZR2019ZD06. 
\end{acks}

\newpage

\bibliographystyle{ACM-Reference-Format}

\end{document}